\pdfoutput=1

\documentclass[11pt]{article}

\usepackage{acl}

\usepackage{appendix}
\usepackage{booktabs,multicol}
\usepackage[utf8]{inputenc} 
\usepackage[T1]{fontenc}    
\usepackage{hyperref}       
\usepackage{url}            
\usepackage{booktabs}       
\usepackage{amsfonts}       
\usepackage{nicefrac}       
\usepackage{microtype}      
\usepackage{lipsum}		
\usepackage{graphicx}
\usepackage{natbib}
\usepackage{doi}
\usepackage{lipsum}  
\usepackage{array}   
\usepackage{booktabs} 
\usepackage[utf8]{inputenc} 
\usepackage[T1]{fontenc}    
\usepackage{hyperref}       
\usepackage{url}            
\usepackage{booktabs}       
\usepackage{amsfonts}       
\usepackage{nicefrac}       
\usepackage{microtype}      
\usepackage{xcolor}
\usepackage{algorithm} 
\usepackage{amsmath}
\usepackage{algpseudocode} 
\usepackage{graphicx}
\usepackage{graphicx}
\usepackage{subcaption}
\usepackage[mathscr]{euscript}
\usepackage{amsmath}
\usepackage{multirow}
\usepackage{booktabs}
\usepackage{times}
\usepackage{latexsym}

\usepackage[T1]{fontenc}

\usepackage[utf8]{inputenc}

\usepackage{microtype}

\usepackage{inconsolata}

\title{RS-DPO: A Hybrid Rejection Sampling and Direct Preference Optimization Method for Alignment of Large Language Models}

\author{ Saeed Khaki\\
	Amazon\\
	\texttt{sakhaki@amazon.com} \\ 
	\And
	JinJin Li\\
	Amazon \\
	\texttt{jinjinli@amazon.com} \\
 	\AND
	Lan Ma\\
	Amazon \\
	\texttt{mamlm@amazon.com}  \\
        \And
        Liu Yang\\
	Amazon \\
	\texttt{liuyanga@amazon.com} \\
        \And
      Prathap Ramachandra\\
	Amazon \\
	\texttt{prathara@amazon.com} \\
}

\begin{document}
\maketitle
\begin{abstract}
Reinforcement learning from human feedback (RLHF) has been extensively employed to align large language models with user intent. However, proximal policy optimization (PPO) based RLHF is occasionally unstable requiring significant hyperparameter finetuning, and computationally expensive to maximize the estimated reward during alignment. Recently, direct preference optimization (DPO) is proposed to address those challenges. However, DPO often relies on contrastive responses generated from human annotator and alternative LLM, instead of the policy model, limiting the effectiveness of the RLHF. In this paper, we addresses both challenges by systematically combining rejection sampling (RS) and DPO. Our proposed method, RS-DPO, initiates with the development of a supervised fine-tuned policy model (SFT). A varied set of k responses per prompt are sampled directly from the SFT model. RS-DPO identifies pairs of contrastive samples based on their reward distribution. Finally, we apply DPO with the contrastive samples to align the model to human preference. Our experiments indicate that our proposed method effectively fine-tunes LLMs with limited resource environments, leading to improved alignment with user intent. Furthermore, it outperforms existing methods, including RS, PPO, and DPO.
\end{abstract}

\section{Introduction}

State-of-the-art (SOTA) LLMs such as GPT-4 \citep{OpenAI2023GPT4TR}, LLaMa \citep{touvron2023llama} etc., are trained with several stages. With pre-training and supervised instruction tuning, LLMs learns to follow specific instructions to complete various tasks with zero-shot or few-shot prompts \citep{chowdhery2022palm}. To further improve the LLMs’ intelligence as close as to human and ensure a more helpful and harmless model, alignment is important as the last-mile LLM training procedure \citep{ziegler2019openairlhf, stiennon2020learning, bai2022training, ouyang2022training}. Reinforcement learning with human feedback (RLHF) \citep{christiano2017deep} is the most adopted approach for alignment training and it usually involves training a reward model with human preference datasets which optimizes a reward function based on the human-annotated preference. Then LLMs are fine-tuned to learn to maximize the reward of their responses using reinforcement learning algorithms, including proximal policy optimization (PPO) \citep{schulman2017proximal}, REINFORCE \citep{Williams2004reinforce}, and similar variants. While PPO is used by SOTA LLMs due to its ease of use and good performance, training with PPO has few limitations, including complexity of training multiple LLMs, and sampling from policy model in training loop, high GPU memory requirement with hosting multiple LLMs during training, and sensitivity to training data and reward models.

To make RLHF training more efficient, there are methods proposed from different perspective. In order to reduce the preference data effort by human annotation, \citep{lee2023rlaif} and \citep{tunstall2023zephyr} proposed to train the LLM to align to the LLM’s preference rating in order to save human effort. \citep{santacroce2023efficient} proposed a combined strategy to merge SFT and reward models as well as in PPO with LoRA selection in order to reduce latency and memory footprint. \citep{dong2023raft, gulcehre2023rest} used reward model to select ranked high-reward good samples to supervise fine-tune the models and iteratively repeating this process yield good results. To reduce the memory and save training resources, \citep{rafailov2023direct} proposed the direct preference optimization (DPO) to remove the need of training reward model, and directly optimize the policy model using a simple classification to maximize the difference between likelihood of human preference pairs. This method proves equivalent performance by implicitly maximize the reward. However, it is mainly trained on human preference data to learn the alignment, instead of sampling the policy model’s response for optimization. LLaMa2 \citep{touvron2023llama} adopts several rounds of rejection sampling to select the best samples from $k$ model-generated samples for fine-tuning before PPO in order to boost the model performance. But rejection sampling only selects the best samples instead of preference pairs, with low data usage efficiency. RSO \citep{liu2023statistical} proposes to generate preference data from the target optimal policy using rejection sampling, enabling a more accurate estimation of the optimal policy. Compared to RSO, our proposed method (RS-DPO) directly employs a point-wise reward model for response ranking and optimization, utilizing logistic loss exclusively during policy optimization. Unlike RSO's approach of statistical rejection sampling and tournament ranking for response generation and selection, RS-DPO generates a fixed number of responses per prompt and relies on computing reward gaps between responses for preference data generation, resulting in reduced computational expense. Additionally, while RSO lacks evaluation on standard alignment benchmarks and comparison against PPO, RS-DPO demonstrates its effectiveness against other RLHF methods on such benchmarks.

In this work, we propose RS-DPO method for RLHF training that combines the advantages of existing efficient methods, including offline preference data generation using rejection sampling, and using DPO in order to reduce the training GPU memory consumption. Specifically, RS-DPO generates responses from the large language model directly, and leverages rejection sampling (RS) to sample synthetic preference pairs based on the reward distribution of LLMs responses. Then, it uses the generated preference pairs for alignment with DPO. The main contributions of our proposed RLHF training method can be summarized as follows: (1) RS-DPO demonstrates stability and robustness against variations in the reward model quality, consistently outperforming existing methods like DPO, PPO and RS. (2) In contrast to the rejection sampling approach that focuses solely on the best response among $k$ generated responses for alignment, RS-DPO selects pairs of contrastive samples based the reward distribution, thereby enhancing overall performance. (3) RS-DPO samples contrastive data directly from the SFT model, distinguishing itself from DPO which often relies on responses from alternative language models or human annotations. This approach contributes to the superior performance of RS-DPO. (4) Our proposed method is efficient, being less resource-intensive compared to PPO, making it practical for applications in limited resource environments.

\section{Method}

The aim of this study is to utilize reinforcement learning from human feedback (RLHF) to train a policy model with the purpose of aligning a large language model to user intent. As the pipeline shown in \ref{fig: whole_flowchart}, our proposed method, RS-DPO, systematically combines RS and DPO. It starts by generating a diverse set of $k$ distinct responses for each prompt, selecting a pair of contrasting samples based on their reward distribution. Subsequently, the method employs DPO to enhance the performance of the language model (LLM), thereby achieving improved alignment. Our proposed method consists the following steps:

\begin{figure*}[h]
    \centering
    \includegraphics[scale=0.3]{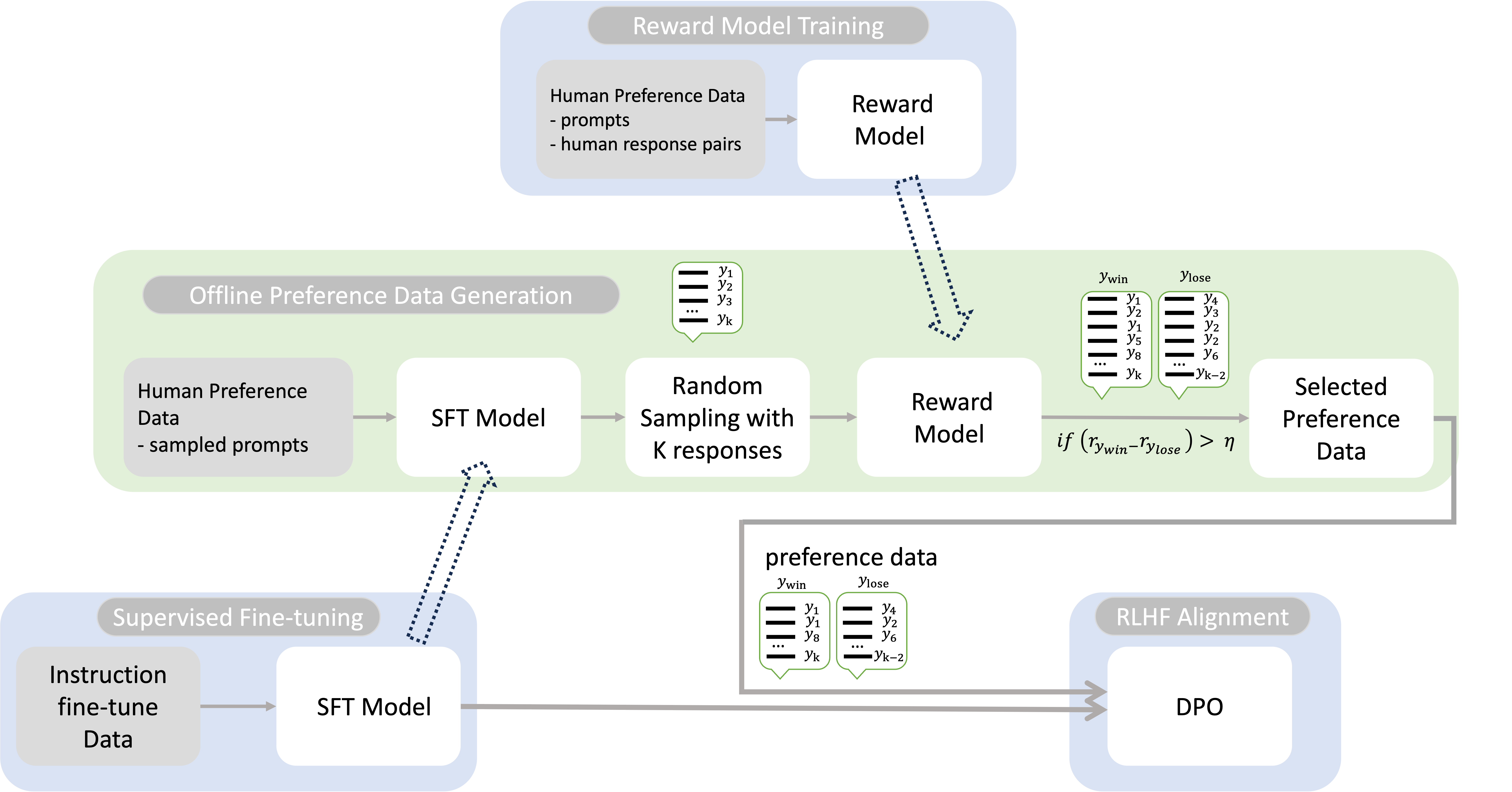}
    \caption{The pipeline of RS-DPO. Our proposed method systematically combines rejection sampling (RS) and direct preference optimization (DPO). We start by creating a SFT model and use it to generate a diverse set of $k$ distinct responses for each prompt. Then, it selects a pair of contrastive samples based on their reward distribution. Subsequently, the method employs DPO to enhance the performance of the language model (LLM), thereby achieving improved alignment.}
    \label{fig: whole_flowchart}
\end{figure*}

\subsection{Supervised Fine-Tuning (SFT)}

As a prerequisite to RLHF, this step involves fine-tuning a pre-trained LLM, $\pi$, using a dataset consisting of high-quality instruction and response pairs or chat data, denoted as $\mathcal{D_{\text{sft}}} = \{(x_1, y_1), \ldots, (x_m, y_m)\}$
\citep{ouyang2022training,wang2023aligning,chung2022scaling,wang2022self}. Starting from a base LLM $\pi$, SFT maximizes the likelihood of response $y$ given prompt $x$ as defined in the Equation \ref{eq:sft}.

\begin{equation}
\mathcal{L^{\text{SFT}}}={\text{argmax}} \sum_{(x,y) \in \mathcal{D_{\text{sft}}}} \log \pi(y | x)\label{eq:sft}
\end{equation}

\subsection{Reward Model Training (RM)}

This step involves training a reward model to assess the quality of a response in accordance with human preferences, with a focus on desired downstream attributes like helpfulness and harmlessness \citep{wang2023aligning,ouyang2022training}. The reward model, denoted as $R(x, y)$, takes a prompt $x$ and a response $y$, and maps them to a scalar value $r$. Let's assume that we have a preference dataset, denoted as $\mathcal{D_{\text{RM}}} = \{(x_1, y_{1l}, y_{1w}), \ldots, (x_n, y_{nl}, y_{nw})\}$, where $x$ represents the input prompt, and $y_l$ and $y_w$ are considered the worse and the better responses, respectively, as determined by human assessment. 
Reward model training uses ranked answers from $\mathcal{D_{\text{RM}}}$ to estimate the preference distribution $p$ as written in Equation \ref{eq:RM} \citep{bradley1952rank}.

\begin{equation}
p(y_w \succ y_l | x) = \frac{\exp{(r(x,y_w))}}{\exp{(r(x,y_w))}+\exp{(r(x,y_l))}}\label{eq:RM}
\end{equation}

Subsequently, we can estimate the parameters of the reward model through maximum likelihood by maximizing the reward gap between $y_w$ and $y_l$, as illustrated in Equation \ref{eq:RM2} \citep{wang2023enable,rafailov2023direct}.


\begin{equation}
\begin{split}
R(x,y) = {\text{argmin}} \sum_{(x,y_l,y_w) \in \mathcal{D}_{\text{RM}}} \\
-\log \sigma(r(x,y_w)) - (r(x,y_l))
\end{split}
\label{eq:RM2}
\end{equation}

\subsection{Preference Data Generation via Rejection Sampling (PDGRS)}\label{sec:PDGRS}
The goal of this step is to create a synthetic preference pair dataset for our alignment task using the trained SFT and RM. Let's denote $\{x_1, \ldots, x_n\}$ as a set of sampled prompts from $\mathcal{D_{\text{RM}}}$ dataset. While it is possible to sample from other prompt datasets, it is crucial to ensure that our prompt sampling remains within the reward model's prompt distribution for optimal performance. We first generate $k$ distinct responses from $\mathcal{L^{\text{SFT}}}$ model for each prompt $x$. Then, we evaluate the quality of each response using our trained reward model $R(x,y)$. Finally, we compute the reward gap for all possible pairwise combinations of responses per prompt, ${k}\choose{2}$. If the reward gap surpasses a predefined threshold, we include the pair of responses in our synthetic preference dataset. The process of preference data generation is illustrated in Algorithm \ref{alg:PDG}.

Since the preference data generation process generates responses from $\mathcal{L_{\text{SFT}}}$ model, it ensures that our RLHF is focused on aligning the $\mathcal{L_{\text{SFT}}}$ behaviour to the human preference rather than distilling knowledge from a larger model or human annotations. We term this process preference data generation via rejection sampling (PDGRS), as it involves evaluating each possible preference data triplet combination (superior and inferior responses), and discarding those with reward gaps below predefined threshold. In addition, our proposed preference data generation process bootstraps and substantially augments the quantity of preference data, compared to the initial static preference dataset $\mathcal{D_{\text{RM}}}$ used in the reward model training.

\begin{algorithm}[h]
\caption{Preference Data Generation via Rejection Sampling}\label{alg:PDG}
\begin{algorithmic}
\State \textbf{Result}:

\State $\mathcal{D_{\text{P}}}=\{(x,y_l,y_w)\}_{3m}$ : Preference dataset  
\State \textbf{Input}:

\State $\{x_1, \ldots, x_n\}$ : Sample prompts from $\mathcal{D_{\text{RM}}}$

\State $\mathcal{L^{\text{SFT}}}$: SFT model

\State $R(x,y)$: Reward model

\State $\tau$: Temperature

\State $\eta$: Threshold for preference data selection


\For{$i=1:n$}
    \State $\,(y_{i1}, \ldots,y_{ik})\,|\,y_{ik}\sim \mathcal{L^\text{SFT}}(\raisebox{0.05ex}{\(\cdot\)}|x_i)$ \Comment{generate $k$ responses from $\mathcal{L^{\text{SFT}}}$ model for prompt $x_i$}

    \State  $\,(r_{i1}, \ldots,r_{ik})\,|\,r_{ij}=R(x_i,y_{ij})$ \Comment{ compute the reward for each of generated responses}

    \For{$j=1:k$}
            \For{$l=1:k$}
                \If{$j==l$}
               \State  continue
                \EndIf
               \State $r_\text{gap}=\,\sigma(\frac{r_{ij}-r_{il}}{\tau})$ \Comment{ compute the reward gap between the pair of responses $y_{il}$ and $y_{ij}$}
                 \If{$r_\text{gap}>\eta$}
                    \State $\mathcal{D_{\text{P}}}=\{\mathcal{D_{\text{P}}};(x_i,y_{il},y_{ij})\}$ \Comment{ append the accepted sample} 
         
                \EndIf
               
            \EndFor
    \EndFor

\EndFor
\end{algorithmic}
\end{algorithm}

\subsection{Direct Preference Optimization (DPO)}

DPO fine-tunes $\mathcal{L^\text{SFT}}$ by directly optimizing the policy model on static preference data $(x, y_l, y_w)$, maximizing the likelihood of the preferred $y_w$ over $y_l$. This approach eliminates the necessity of fitting an explicit reward model by using the ratio of likelihood between the policy $\mathcal{L^\text{RL}}$ model and the original $\mathcal{L^{\text{SFT}}}$ model as an implicit reward signal \citep{rafailov2023direct}. During training process, DPO optimizes the objective function as written in Equation \ref{eq:DPO}.

\begin{equation}
\begin{split}
\mathcal{L}^\text{RL} = {\text{argmax}} &\sum_{(x,y_l,y_w) \in \mathcal{D}_{\text{P}}} \\
&\log \sigma(\beta\log\frac{\mathcal{L}^\text{RL}(y_w|x)}{\mathcal{L}^\text{SFT}(y_w|x)} \\
&-\beta\log\frac{\mathcal{L}^\text{RL}(y_l|x)}{\mathcal{L}^\text{SFT}(y_l|x)})
\end{split}
\label{eq:DPO}
\end{equation}


Our proposed method employs DPO on the synthetic preference dataset $\mathcal{D_{\text{P}}}$ generated in step \ref{sec:PDGRS} using PDGRS, in order to align the policy model with human preferences.

\section{Experiments Details}

This section outlines our experiments to demonstrate the effectiveness of our proposed RS-DPO method for the alignment task. We conduct all of our experiments on the Llama-2-7B LLM \citep{touvron2023llama} which is one of the state-of-the-art LLMs at 7B parameter scale. We perform supervised fine-tuning (SFT), reward modeling (RM), DPO training, and PPO training based on Huggingface Transformer Reinforcement Learning (TRL) library \citep{vonwerra2022trl}. We utilize DeepSpeed ZeRO-3 \citep{rajbhandari2020zero} for optimizing GPU memory and training speed. All experiments are conducted on 8 A100s GPUs with 40G memory per GPU. 

\subsection{Datasets}

We use the following datasets in our experiments:


\textbf{Open Assistant:} Open Assistant (OASST1) \citep{kopf2023openassistant} is a multilingual human-generated conversation dataset ranked for quality. In our experiment, we  utilize the highest quality partition based on quality ranking, comprising of 9k samples.

\textbf{Anthropic/HH-RLHF:} Anthropic released this dataset that includes 169.55k conversation pairs between humans and an AI assistant to train a helpful and safe AI assistant. This preference dataset has two subsets namely helpfulness and harmlessness \citep{bai2022training,ganguli2022red}. In our experiments, we only use a random sample of the helpfulness subset of the data with the size of roughly 10,300 samples.

\textbf{WebGPT:} WebGPT \citep{nakano2021webgpt} dataset includes long-form question answering preference dataset annotated by humans for reward modeling. After cleaning this dataset, we get 17,814 samples from this dataset.

\subsection{Experimental Setup}\label{sec: ex_setup}

We start our experiments by training a Llama-2-7B SFT model using the Open Assistant conversation dataset. We specifically choose this SFT dataset for two primary reasons: (1) the same SFT model is used across different preference datasets in RLHF. This helps to examine the influence of preference dataset on our proposed method, and (2) the utilization of high-quality chat data leads to the improved performance of SFT models \citep{dettmers2023qlora}. For SFT step, we employ linear learning rate schedule with starting learning rate of $2\times 10^{-5}$, effective batch size of 64, number of epochs of 2, weight decay of 0.1, and a sequence length of 4096 tokens. We do not use LoRA \citep{hu2021lora} finetuning in the SFT step.

In the response generation step during the PDGRS (\ref{alg:PDG}) of our proposed method, we generate a total of $k=16$ responses for each prompt, with the following decoding parameters: a maximum of 512 new tokens, a $\text{top-k}$ value of 50, a $\text{top-p}$ value of 0.98, and a sampling temperature of 1. We applied PDGRS on 10,300 samples from Anthropic/HH-RLHF, and 12,193 samples from WebGPT.

To assess the quality of generated responses, we employ the pythia-6.9B reward model developed by Open Assistant, denoted as pythia-6.9B-RM-OA in our experiments \citep{openassistantPythia}. This reward model is trained on a diverse set of datasets, including Open Assistant preference \citep{kopf2023openassistant}, Anthropic \citep{bai2022training,ganguli2022red}, SHP \citep{pmlr-v162-ethayarajh22a}, hellaswag \citep{zellers2019hellaswag}, WebGPT \citep{nakano2021webgpt}, and summary pairs \citep{stienon2020learning}. To control the impact of reward model preference data with our proposed method, we also trained a pythia-6.9B reward model using only WebGPT preference dataset, denoted as pythia-6.9B-RM-WG in our experiments. We use pythia-6.9B \citep{biderman2023pythia} as a base model and train it for 1 epoch with learning rate of $1\times 10^{-5}$ with linear learning rate schedule.

For DPO training in our experiments, we use cosine learning rate schedule with an initial learning rate of $1\times 10^{-6}$, effective batch size of 64, number of epochs of 4, $\beta=0.1$, and a sequence length of 4096 tokens. We use LoRA with $\text{rank}=8$ to enable training Llama-2-7B models with limited GPU resources.

For PPO \citep{schulman2017proximal} training in our experiments, we use LoRA with $\text{rank}=8$ and 8-bit quantization for both policy and reward models. We adopt effective batch size of 64, learning rate of $2\times 10^{-5}$, and Kullback-Leibler (KL) coefficient of 0.2. We train the policy model between 150-200 steps to converge.

\subsection{Evaluation}

Assessing alignment to human preference is challenging, but recent developments have introduced specialized benchmarks like MT-Bench \citep{zheng2023judging} and AlpacaEval \citep{alpaca_eval} to address this issue. These benchmarks leverage strong LLM judges like GPT-4, providing a score that strongly correlates with human preference ratings. We use the following benchmarks to evaluate model's performance on instruction following and alignment to user intent:

\textbf{MT-Bench:} MT-Bench evaluation is based on GPT-4 judgement and achieves over 80\% agreement with human preference. MT-bench is designed to test multi-turn conversation and instruction-following ability of LLMs, covering 8 common categories including writing, roleplay, extraction, reasoning, math, coding, knowledge I (STEM), and knowledge II (humanities/social science). MT-Bench has 10 multi-turn questions for each category, and GPT-4 rates each turn's response on a scale of 1-10, with the final score being the mean over two turns \citep{zheng2023judging}. 

\textbf{AlpacaEval:} It is an LLM-based automatic evaluation judged by GPT-4, where it measures the pairwise win-rate against a baseline model (text-davinci-003). We use 300 questions mostly focused on helpfulness from this benchmark in our evaluations \citep{alpaca_eval}.

\section{Results and Ablations}
This section presents our main results. We show sample model completions in appendix \ref{Appendix:C}. To comprehensively assess the effectiveness of our proposed method, we employ a comparative analysis of various preference data generation policies. These policies guide the selection of the superior model response, denoted as $y_w$, and the inferior model response, denoated as $y_l$, from a set of $k$ generated answers. The following preference data generation policies are considered:

\textbf{Best-vs-worst:} This policy ranks the $k$ responses according to their respective rewards and selects the response with the highest reward as $y_w$ and the response with the lowest reward as $y_l$.

\textbf{Best-vs-random:} This policy selects the response with the highest reward as $y_w$, while $y_l$ is chosen randomly from the remaining $k-1$ responses.

\textbf{Original annotation:} This policy chooses $y_w$ and $y_l$ from the  original preference data annotated by humans or larger models.

\textbf{Rejection Sampling:} This method utilizes only the response with the highest reward as $y_w$ for each prompt $x$ and performs 1-step SFT using samples $(x,y_w)$.

\textbf{PPO:} This method dynamically generates responses $y$ for a batch of prompts and employs a reward model for their assessment. Subsequently, it maximizes the cumulative reward during RLHF training. PPO does not use any pre-generated responses.

\textbf{Proposed method:} We consider all possible combinations of $y_w$ and $y_l$ from the $k$ answers. We keep all combinations with reward gap larger than predefined threshold $\eta$.

We also investigate the performance of our proposed method under varying thresholds $\eta$, maintaining a constant temperature $\tau$. Generally, lower values of $\eta$ and $\tau$ lead to an increased size of preference data in our proposed method. In contrast to other policies that limit the sample size to the original preference data $\mathcal{D_{\text{RM}}}$ size, our proposed method considers the reward distribution per prompt and identifies more contrastive samples $(y_w,y_l)$, thereby resulting in enhanced performance. Table \ref{tab:performance_anthropic} and Table \ref{tab:performance_webgpt} summarizes the results on the MT-Bench and AlpacaEval benchmarks for Anthropic/HH-RLHF and WebGPT datasets, respectively.

\begin{table*}[h]
    \centering
    \captionsetup{skip=6pt} 
      \resizebox{\textwidth}{!}{
    \begin{tabular}{cccc|cc}
        \toprule
        \textbf{Policy} &\textbf{Reward Model} & \textbf{Sample Size}&\textbf{Threshold}   & \textbf{MT-Bench (Avg score)} & \textbf{ AlpacaEval (win \%)} \\
        \midrule
        SFT & - &9,000 & -& 5.12&$60.20_{2.84}$  \\
         Best-vs-worst & pythia-6.9B-RM-OA & 10,300 &- &  5.34&$72.48_{2.59}$ \\
         Best-vs-random & pythia-6.9B-RM-OA &10,300 &- &  5.07 &$70.00_{2.64}$ \\
         Original annotation & - &10,300 & - &  5.26 &$65.33 _{2.75}$\\
        Rejection Sampling & pythia-6.9B-RM-OA &10,300 & - &  4.84&$60.20_{2.84}$ \\
        PPO & pythia-6.9B-RM-OA &10,300 & - &  5.22&$69.23_{2.67}$ \\
         Proposed method & pythia-6.9B-RM-OA &12,795 & 0.90 & 5.44&$73.75 _{2.54}$ \\
      Proposed method & pythia-6.9B-RM-OA &32,640 & 0.85 & $\mathbf{5.49}$&$74.17_{2.53}$ \\
         Proposed method & pythia-6.9B-RM-OA &63,938 & 0.80 & 5.36& $\mathbf{79.67_{2.33}}$\\
        \bottomrule
    \end{tabular}
    }
     \caption{Performance of competing methods on Anthropic/HH-RLHF dataset using different policies on MT-Bench and AlpacaEval benchmarks. A dash (-) sign indicates that the specific parameters is not needed. The SFT model is trained on Open Assistant conversation dataset. The base LLM for all experiments is  Llama-2-7B. The temperature $\tau$ is set to be 1 in our proposed method. The subscript in the AlpacaEval win rate indicates the standard error.}
    \label{tab:performance_anthropic}
\end{table*}


\begin{table*}[h]
    \centering
    \captionsetup{skip=6pt} 
          \resizebox{\textwidth}{!}{
    \begin{tabular}{cccc|cc}
        \toprule
      \textbf{Policy} &\textbf{Reward Model} & \textbf{Sample Size}&\textbf{Threshold}   & \textbf{MT-Bench (Avg score)} & \textbf{ AlpacaEval (win \%)} \\
        \midrule
        SFT & - &9,000 & - & 5.12 &$60.20_{2.84}$\\
         Best-vs-worst & pythia-6.9B-RM-WG & 12,193 & - &  5.24 &$69.17_{2.67}$\\
         Best-vs-random & pythia-6.9B-RM-WG &12,193 & - & 5.04 &$69.90_{2.66}$\\
         Original annotation & - &12,193 & - &  5.14 &$65.55_{2.75}$\\
        Rejection Sampling & pythia-6.9B-RM-WG &12,193 & - & 5.15 &$68.17_{2.69}$\\
        PPO & pythia-6.9B-RM-WG &12,193 & - & 4.95 &$65.17_{2.75}$\\
         Proposed method & pythia-6.9B-RM-WG &3,449 & 0.90 &  5.13 &$68.90_{2.68}$\\
      Proposed method & pythia-6.9B-RM-WG &11,458 & 0.85 &  5.24 &$72.33_{2.59}$\\
         Proposed method & pythia-6.9B-RM-WG &29,698 & 0.80 &  5.31 &$72.91_{2.57}$\\
         Rejection Sampling & pythia-6.9B-RM-OA &12,193 & - &  5.23 &$71.00_{2.62}$\\
          PPO & pythia-6.9B-RM-OA &12,193 & - &  5.11 &$69.83_{2.65}$\\
         Proposed method & pythia-6.9B-RM-OA &12,611 & 0.90 &$\mathbf{5.35}$&$71.91_{2.60}$\\
      Proposed method & pythia-6.9B-RM-OA &33,755 & 0.85 &  $\mathbf{5.35}$ &$\mathbf{74.92_{2.51}}$\\
         Proposed method & pythia-6.9B-RM-OA &70,510 & 0.80 &  5.20& $67.56_{2.71}$\\
        \bottomrule
    \end{tabular}
    }
     \caption{Performance of competing methods on WebGPT dataset using different policies on MT-Bench and AlpacaEval benchmarks. A dash (-) sign indicates that the specific parameters is not needed. The SFT model is trained on Open Assistant conversation dataset. The base LLM for all experiments is  Llama-2-7B. The temperature $\tau$ is set to be 1 in our proposed method. The subscript in the AlpacaEval win rate indicates the standard error.}
    \label{tab:performance_webgpt}
\end{table*}

In Table \ref{tab:performance_anthropic} and Table \ref{tab:performance_webgpt}, our proposed method consistently demonstrates superior performance compared to other methods on the Anthropic/HH-RLHF and WebGPT datasets. All policies exhibit better performance than the SFT model, except for the best-vs-random policy and PPO on MT-Bench benchmark. This can be attributed to the best-vs-random policy's random selection of $y_l$, which, if it happens to select a high-quality response as $y_l$, can make optimization process challenging and noisy. 

The best-vs-worst policy consistently outperforms other policies except our proposed method, primarily because it consistently selects high-quality pairs of contrastive samples. Furthermore, the best-vs-worst policy also outperforms the original annotation policy, despite both policies utilizing the same amount of data. This observation holds true even for pythia-6.9B-RM-WG, which is trained on the same original annotation dataset. The enhanced performance of the best-vs-worst policy can be attributed to the fact that both $y_l$ and $y_w$ are sampled from the SFT model, as opposed to utilizing responses from another language model or human annotation. 

Rejection sampling method is not performing very well which can be attributed the following factors: (1) it only utilizes $y_w$ for alignment and does not take advantage of the remaining $k-1$ responses, (2) it applies 1-step SFT which can be susceptible to overfitting issues.

The performance of PPO on Anthropic/HH-RLHF surpasses that of other methods, with the exception of our proposed approach and the best-vs-worst policy. However, the performance of PPO on MT-Bench average scores declines when applied to WebGPT, primarily attributed to a low 2-turn score on MT-Bench, as detailed in Tables \ref{tab:performance_anthropic-full} and \ref{tab:performance_webgpt-full} in appendix. This can be attributed to the prompt types in the datasets, where the Anthropic/HH-RLHF dataset comprises prompts featuring multi-turn conversations between humans and AI assistants, while the WebGPT dataset exclusively involves single-turn questions. Consequently, PPO indicates an enhancement in second-turn performance on the Anthropic/HH-RLHF dataset in comparison to WebGPT within the MT-Bench benchmark.

\textbf{How does changing the threshold $\eta$ affect our performance of proposed method?} 
Our proposed method takes into account the reward distribution per prompt to determine pairs of $y_l$ and $y_w$ by assessing the reward gap. Lower values of $\eta$ lead to an increased generation of preference data within our proposed method because it allows selection of samples with smaller reward gaps. However, setting $\eta$ too low may lead to $y_l$ and $y_w$ being similar in quality, potentially impeding the optimization process and the convergence. In both datasets, reducing $\eta$ from 0.90 to 0.85 yields improved performance as it increases preference data generation without compromising quality. However, lowering $\eta$ further, from 0.85 to 0.80, results in a slight performance decline in two cases when using the pythia-6.9B-RM-OA reward model on MT-Bench bench. This can be attributed to a substantial increase in sample size, preventing the convergence of the optimization process and reduced quality of generated preference data.

\begin{figure*}[h]
    \centering
    \includegraphics[scale=0.3]{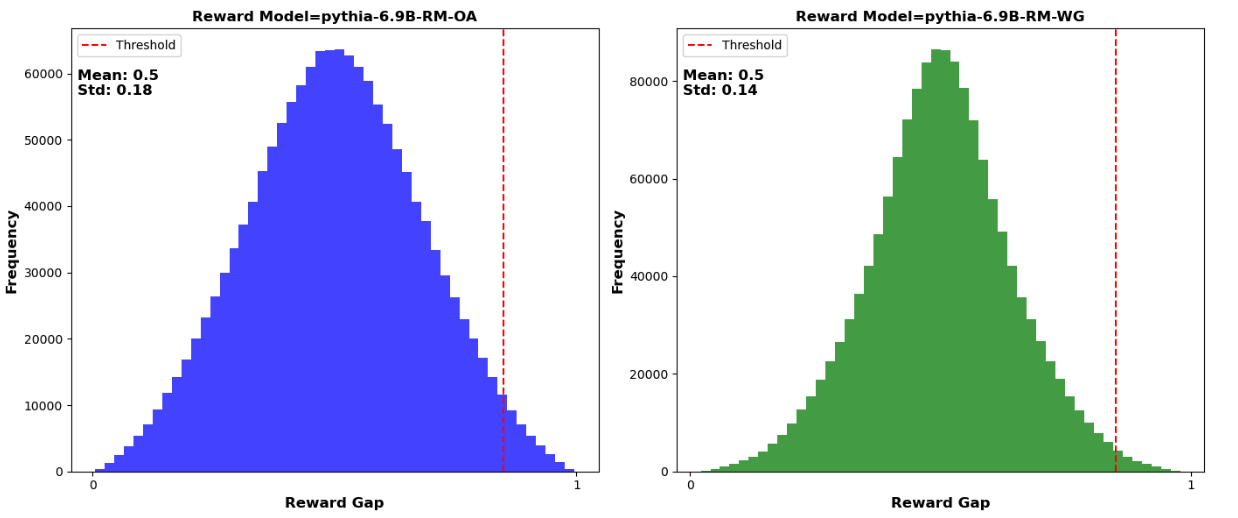}
    \caption{Histograms of reward gap for WebGPT datasets with different reward models. The red dashed line represents the threshold value of 0.85 for preference data selection. Mean and standard deviation values of reward gaps are shown in the histograms.}
    \label{fig:reward_gap_webgpt}
\end{figure*}

\textbf{How does the reward model impact the results?} In our experiments, we employ two reward models with identical architectures but trained on different amount of preference data. Specifically, pythia-6.9B-RM-OA is trained on a larger preference dataset, while pythia-6.9B-RM-WG is exclusively trained on the WebGPT portion of preference datasets (detailed information is provided in section \ref{sec: ex_setup}). As a result, pythia-6.9B-RM-OA exhibits superior performance in evaluating response quality in line with human preferences. Typically, a more effective reward model tends to have a higher variance in its reward distribution with longer tails, as it can differentiate the good and bad responses in a broader range. In contrast, lower quality reward models often have most rewards concentrated around the mean. Figure \ref{fig:reward_gap_webgpt} shows the reward gap distribution for both reward models on WebGPT dataset. The red dashed line represents the threshold for preference data selection in the histograms. As depicted in Figure \ref{fig:reward_gap_webgpt}, the histogram for pythia-6.9B-RM-OA exhibits longer tails and greater variance, leading to a higher number of preference samples falling in after the dashed line. As shown in the reward model ablation study in Table \ref{tab:performance_webgpt}, the pythia-6.9B-RM-OA reward model enhances the performance of our proposed method, PPO, and the rejection sampling method, underscoring the significance of a high-quality reward model. Nevertheless, the results demonstrate the robustness of our proposed method to variations in reward model quality, as it outperforms other methods even when employing the pythia-6.9B-RM-WG reward model. Additionally, results indicates that the PPO method is more sensitive to the quality of the reward model as transitioning from the pythia-6.9B-RM-OA to the pythia-6.9B-RM-WG reward model notably diminishes model performance across both benchmarks.

\textbf{How do multi-turn prompts influence performance?}

The Anthropic/HH-RLHF dataset includes prompts comprising multi-turn conversations between humans and AI assistants, while the WebGPT dataset exclusively consists of single-turn questions. Through a comparative analysis of MT-bench multi-turn scores presented in Table \ref{tab:performance_anthropic-full} and \ref{tab:performance_webgpt-full}, it is evident that the incorporation of multi-turn prompts enhances the 2-turn scores for both our proposed method and PPO. Notably, our proposed method outperforms PPO. However, there is no significant impact on performance observed in the AlpacaEval benchmark as it employs only single-turn evaluation prompts. Consequently, the inclusion of multi-turn prompts in RLHF is crucial for improving the multi-turn capabilities of language models.

\textbf{How does changing the temperature $\tau$ affect our method's performance?} To analyze the impact of temperature on our proposed method, we design an ablation study where we keep the threshold $\eta=0.85$ the same and change the value of temperature. Decreasing $\tau$ leads to a heavy-tailed reward gap distribution, generating more preference samples. Conversely, increasing $\tau$ creates a thin-tailed reward gap distribution, resulting in fewer preference samples being generated. We conduct this ablation study on Anthropic/HH-RLHF dataset and Table \ref{tab:performance_anthropic_temp} summarizes the results. A lower temperature value increases the sample size, enhancing overall performance. As the sample size increases, a diverse variety of preference pairs emerges, encompassing both easy (with a higher reward gap) and hard (with a lower reward gap) instances. The inclusion of easy preference pairs significantly helps with the convergence of the DPO optimization process, facilitating the attainment of a superior model. In contrast, solely using hard samples may impede the optimization process, resulting in a failure to converge and yielding a policy model of inferior quality.

\begin{table*}[b]
    \centering
    \captionsetup{skip=6pt} 
          \resizebox{\textwidth}{!}{
    \begin{tabular}{cccc|cc}
        \toprule
        \textbf{Policy}  & \textbf{Sample Size}&\textbf{Threshold} & \textbf{Temperature} & \textbf{MT-Bench (score)} & \textbf{ AlpacaEval (win \%)}  \\
        \midrule
 
         Proposed method  &63,796 & 0.85 & 0.8  & 5.31 &$\mathbf{77.33_{2.42}}$\\
      Proposed method  &45,668 & 0.85 & 0.9 &  $\mathbf{5.51}$ &$76.92_{2.44}$\\
          Proposed method  &32,640 & 0.85 & 1 &  5.49 &$74.17_{2.53}$\\
             Proposed method  &22,951 & 0.85 & 1.1 &  5.40 &$71.00_{2.62}$\\

                  Proposed method  &16,160 & 0.85 & 1.2 &  5.43 &$71.33_{2.62}$\\
        \bottomrule
    \end{tabular}
    }
     \caption{Performance of our proposed method on Anthropic/HH-RLHF dataset using different temperature $\tau$ on MT-Bench and AlpacaEval benchmarks. The base LLM for all experiments is  Llama-2-7B. The reward model for all methods is pythia-6.9B-RM-OA. The subscript in the AlpacaEval win rate indicates the standard error.}
    \label{tab:performance_anthropic_temp}
\end{table*}

\section{Discussion and Conclusion}

In this paper, we proposed RS-DPO method that generates responses from the large language model directly, and leverages RS to sample synthetic preference pairs, and DPO for RLHF training. Extensive experiments show the effectiveness of RS-DPO compared to existing methods including rejection sampling (RS), proximal policy optimization (PPO) and direct preference optimization (DPO). Additionally, RS-DPO is stable, and is not as sensitive to the quality of the reward model as other methods. Our proposed method also offers a more efficient and less time-consuming solution for the alignment task as compared to PPO, minimizing resource requirements.

During RLHF training, PPO conducts online sampling from the policy model and evaluates them using the loaded reward model in real-time. Consequently, PPO necessitates loading three models during training: the initial SFT, policy model, and reward model, demanding a significant amount of GPU memory and decelerating the training process. Furthermore, the online sampling from the policy model incurs increased memory consumption as the generated sequences lengthen. In practical terms, even with 1-2 moderate GPUs, training a small-scale (e.g., 7B) LLM using PPO is unfeasible. In our experiments, we had 8 A-100 GPUs each having 40G memory, but we resorted to 8-bit quantization of both the policy and reward model to circumvent GPU memory constraints. Our proposed method conducts response sampling offline from SFT and constructs a dataset of synthetic preference data to bypass the high computational cost of PPO, while remaining viable on 1-2 moderate GPUs. Notably, the operational cost of running DPO and RS-DPO is identical; the sole disparity lies in RS-DPO performing offline SFT sampling, rendering our proposed method an on-policy reinforcement learning approach.

Moreover, as emphasized by prior researches \citep{singhal2023long}, PPO represents an unstable process prone to sensitivity towards reward model quality and hyperparameters, necessitating multiple runs to converge to a satisfactory model. For instance, in Table 2, training two models using PPO while altering the reward model quality from high (pythia-6.9B-RM-OA) to lower (pythia-6.9B-RM-WG) significantly impacted the resulting model's quality, underscoring PPO's sensitivity. Conversely, our proposed method exhibits robustness against reward model quality, requiring only a single run to train each model successfully.

\section{Limitations}
A limitation of our work is its primary focus on the helpfulness objective derived from open-source preference datasets. Consequently, the generalizability of our findings to other objectives, such as harmlessness may be constrained. While we have demonstrated the efficacy of our proposed method on language models at 7B scale, we acknowledge that we have yet to subject our method to larger or close-source models. Despite these limitations, we maintain confidence that our proposed method demonstrates robustness towards reward model quality, and needs fewer resources compared to existing methods of RLHF training.

\bibliography{references}

\appendix



\section{MT-Bench Benchmark Multi-turn Results}\label{sec:appendix_mt_bench}
MT-bench consists of a multi-turn question set which is deigned for testing the multi-turn conversation and instruction-following ability of LLMs. In the section, we present the MT-bench scores for all individual turns in Tables \ref{tab:performance_anthropic-full} and \ref{tab:performance_webgpt-full}.

\begin{table*}[h]
    \centering
    \captionsetup{skip=6pt} 
      \resizebox{\textwidth}{!}{
    \begin{tabular}{cccc|ccc}
        \toprule
        \textbf{Policy} &\textbf{Reward Model} & \textbf{Sample Size}&\textbf{Threshold} &  \textbf{Turn-1} & \textbf{Turn-2} & \textbf{Average} \\
        \midrule
        SFT & - &9,000 & - & 5.70 & 4.54 & 5.12 \\
         Best-vs-worst & pythia-6.9B-RM-OA & 10,300 &- &  6.06 & 4.61 & 5.34 \\
         Best-vs-random & pythia-6.9B-RM-OA &10,300 &- &  5.77 & 4.38 & 5.07 \\
         Original annotation & - &10,300 & - & 5.89 & 4.62 & 5.26 \\
        Rejection Sampling & pythia-6.9B-RM-OA &10,300 & - &  5.54 & 4.13 & 4.84 \\
         PPO & pythia-6.9B-RM-OA &10,300 & - &  6.03 & 4.41 & 5.22 \\
         Proposed method & pythia-6.9B-RM-OA &12,795 & 0.90 &  5.96 & \textbf{4.91} & 5.44 \\
      Proposed method & pythia-6.9B-RM-OA &32,640 & 0.85 &  \textbf{6.18} & 4.81 & \textbf{5.49} \\
         Proposed method & pythia-6.9B-RM-OA &63,938 & 0.80 &  6.07 & 4.63 & 5.36 \\
        \bottomrule
    \end{tabular}
    }
     \caption{Performance of competing methods on Anthropic/HH-RLHF dataset using different policies on MT-Bench benchmark. We report turn-1, turn-2, and average score from MT-Bench judged by GPT-4. A dash (-) sign indicates that the specific parameters is not needed. The SFT model is trained on Open Assistant conversation dataset. The base LLM for all experiments is  Llama-2-7B.}
    \label{tab:performance_anthropic-full}
\end{table*}


\begin{table*}[h]
    \centering
    \captionsetup{skip=6pt} 
      \resizebox{\textwidth}{!}{
    \begin{tabular}{cccc|ccc}
        \toprule
        \textbf{Policy} &\textbf{Reward Model} & \textbf{Sample Size}&\textbf{Threshold}  & \textbf{Turn-1} & \textbf{Turn-2} & \textbf{Average} \\
        \midrule
        SFT & - &9,000 & - &  5.70 & 4.54 & 5.12 \\
         Best-vs-worst & pythia-6.9B-RM-WG & 12,193 & - &  5.85 & 4.63 & 5.24 \\
         Best-vs-random & pythia-6.9B-RM-WG &12,193 & - &  5.61 & 4.45 & 5.04 \\
         Original annotation & - &12,193 & - & 5.80 & 4.48 & 5.14 \\
        Rejection Sampling & pythia-6.9B-RM-WG &12,193 & - &  5.66 & 4.63 & 5.15 \\
          PPO & pythia-6.9B-RM-WG &12,193 & - &  5.64 & 4.26 & 4.95 \\
         Proposed method & pythia-6.9B-RM-WG &3,449 & 0.90 &  5.83& 4.44 & 5.13 \\
      Proposed method & pythia-6.9B-RM-WG &11,458 & 0.85 &  5.86& 4.63 & 5.24 \\
         Proposed method & pythia-6.9B-RM-WG &29,698 & 0.80 &  5.87 & 4.73 & 5.31 \\
         Rejection Sampling & pythia-6.9B-RM-OA &12,193 & - &  5.94 & 4.49 & 5.23 \\
         PPO & pythia-6.9B-RM-OA &12,193 & - &  5.94 & 4.27 & 5.11 \\
         Proposed method & pythia-6.9B-RM-OA &12,611 & 0.90 &  5.91 & \textbf{4.77} & \textbf{5.35} \\
      Proposed method & pythia-6.9B-RM-OA &33,755 & 0.85 &  \textbf{6.13} & 4.55 & \textbf{5.35} \\
         Proposed method & pythia-6.9B-RM-OA &70,510 & 0.80 &  5.96 & 4.42 & 5.20 \\
        \bottomrule
    \end{tabular}
    }
     \caption{Performance of competing methods on WebGPT dataset using different policies on MT-Bench benchmark. We report turn-1, turn-2, and average score from MT-Bench judged by GPT-4. A dash (-) sign indicates that the specific parameters is not needed. The SFT model is trained on Open Assistant conversation dataset. The base LLM for all experiments is  Llama-2-7B.}
    \label{tab:performance_webgpt-full}
\end{table*}

\section{DPO Reward Accuracy and Reward Margin}

\begin{figure*}[h]
    \centering
    \includegraphics[scale=0.4]{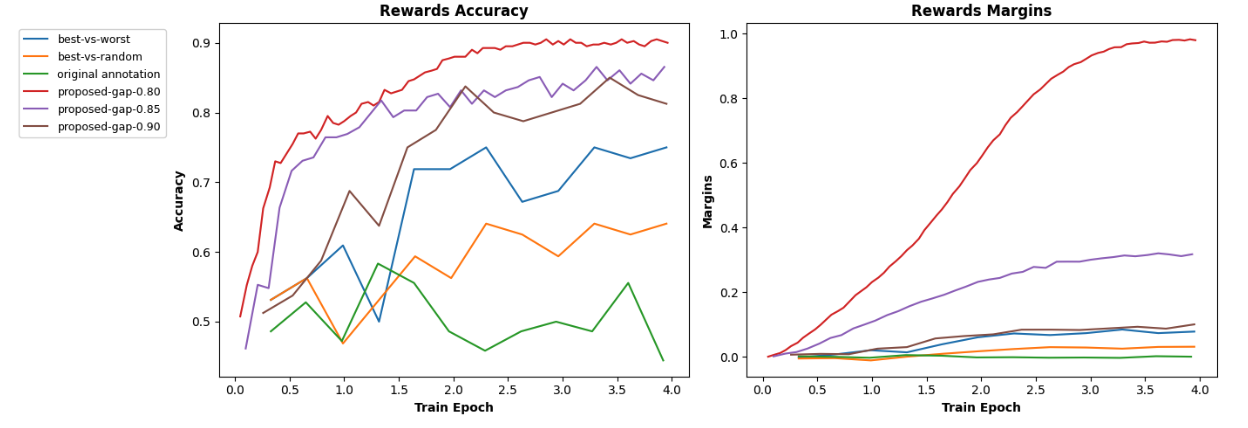}
    \caption{The left and right plots depict the reward accuracy and reward margin, respectively, of competing methods during DPO training on the Anthropic/HH-RLHF dataset. }
    \label{fig:dpo_hhh}
\end{figure*}

\begin{figure*}[h]
    \centering
    \includegraphics[scale=0.34]{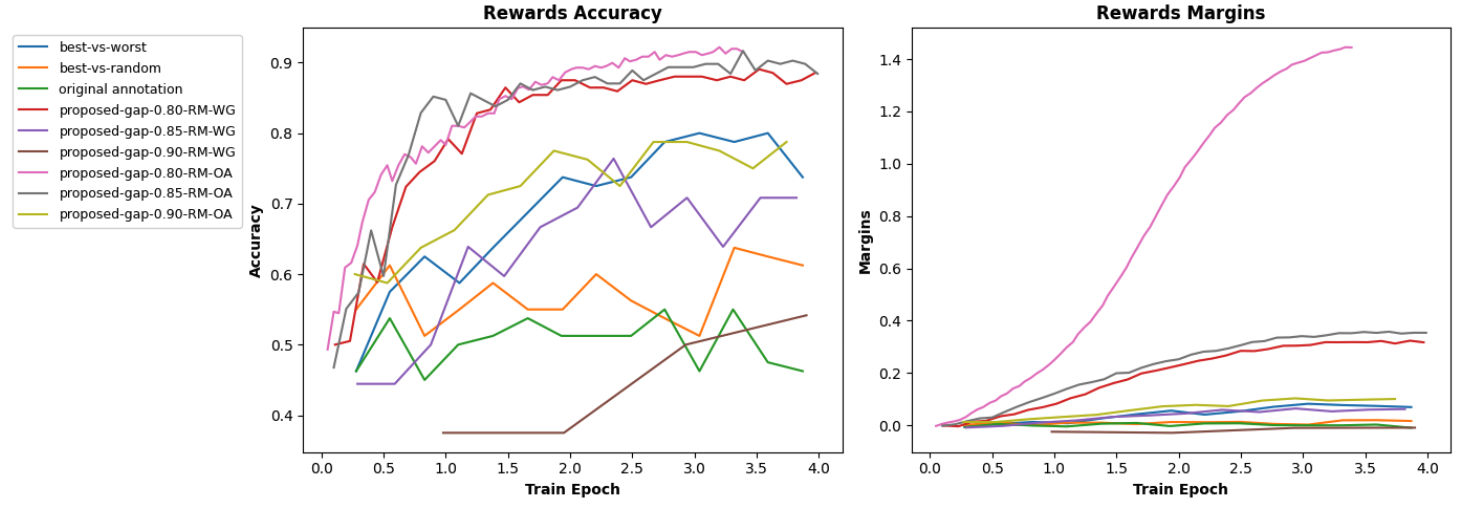}
    \caption{The left and right plots display the reward accuracy and reward margin, respectively, of competing methods during DPO training on the WebGPT dataset.}
    \label{fig:dpo_webgpt}
\end{figure*}

By employing our proposed PDGRS methodology to generate preference datasets, we leverage the DPO method to fine-tune the policy model, enhancing its alignment with human preferences. Figures \ref{fig:dpo_hhh} and \ref{fig:dpo_webgpt} illustrate the reward margins and accuracies achieved through DPO training across various methods on the hold-out evaluation datasets for Anthropic/HH-RLHF and WebGPT, respectively. According to the results, we observe significant correlation between increased reward margins, accuracies, and improved model performance. Our proposed preference data generation method indicates superior reward accuracy and margin in the plots, thereby underscoring the high data quality in our preference data generation approach.

\section{Sample Size Controlling in RS-DPO}\label{Appendix:D}

One advantage of our proposed method is its capacity to generate preference data by considering the reward distribution per prompt. This approach allows us to determine pairs of $y_l$ and $y_w$ by assessing the reward gap, freeing our sample size from being bound to the number of prompts in the data, unlike methods such as DPO or the Best-vs-worst method. Our results demonstrate that increasing the sample size enhances the performance of our proposed method. However, to control for the sample size's effect and showcase our method's performance when the sample size is equivalent to other methods, we subsample the generated preference data from our method to match the original number of prompts, which are 10,300 and 12,193 for the Anthropic/HH-RLHF and WebGPT datasets, respectively. Table \ref{tab:sample_size} provides a summary of the results on the MT-Bench benchmark. Compared to Tables \ref{tab:performance_anthropic} and \ref{tab:performance_webgpt}, our results indicate that controlling the sample size has no impact on the performance of our proposed method, which continues to outperform other methods.

\begin{table*}[h]
    \centering
    \captionsetup{skip=6pt} 
          \resizebox{\textwidth}{!}{
    \begin{tabular}{ccccc|c}
        \toprule
      \textbf{Policy} &\textbf{Reward Model} & \textbf{Sample Size}&\textbf{Threshold}   & \textbf{Dataset} &  \textbf{MT-Bench (Avg score)} \\
        \midrule
   
      Proposed method & pythia-6.9B-RM-OA &10,300 & 0.90 & Anthropic/HH-RLHF  &$\mathbf{5.37}$\\
            Proposed method & pythia-6.9B-RM-OA &12,193 & 0.90 & WebGPT  &$\mathbf{5.35}$\\
        \bottomrule
    \end{tabular}
    }
     \caption{Performance of our proposed method on MT-Bench benchmark with sample size controlling in Anthropic/HH-RLHF and WebGPT datasets.}
    \label{tab:sample_size}
\end{table*}

\section{Qualitative Examples}\label{Appendix:C}

 To conduct a qualitative comparison of model responses trained through various methods, we select sample prompts from two benchmark datasets, namely MT-Bench and AlpacaEval. Subsequently, responses are generated across all candidate models. The results of this comparative analysis are presented in Tables \ref{tab:sample_mt_bench},  \ref{tab:sample_alpaca_eval2} and \ref{tab:sample_alpaca_eval}.
\begin{table*}[h]
  \centering
  \renewcommand{\arraystretch}{1.5}
  \small
  \begin{tabular}{cp{13cm}}
    \hline
    \multicolumn{2}{p{16cm}}{\textbf{Prompt:} You can see a beautiful red house to your left and a hypnotic greenhouse to your right, an attractive heated pink place in the front. So, where is the White House?} \\
    \hline
    Policy & Response \\
    \hline
    Proposed Method & The White House is the official residence and workplace of the president of the United States. It is located at 1600 Pennsylvania Avenue NW in Washington, D.C. It is a complex of buildings that includes the West Wing, the East Wing, the Residence, and the Eisenhower Executive Office Building. The White House is often referred to as "the People's House" because it is the symbol of the American people and the presidency. It is a historic landmark and a popular tourist destination. The red house, greenhouse, and heated pink place are not located near the White House, but they are likely other attractions or landmarks in the area. The White House can be easily recognized by its iconic white exterior and its prominent location in Washington, D.C. \\
     \hline
    SFT & The White House is the large, white building in the center of the picture. It is located in Washington, D.C., and is the official residence and workplace of the president of the United States. The red house to the left is the Blair House, which is used for official guest accommodations. The greenhouse to the right is the United States Botanic Garden, which is a research and educational institution. The attractive heated pink place in the front is the Ellipse, a park located between the White House and the Washington Monument. \\

    \hline
  Original Annotation &The White House is the building in the center of the picture. It is the building with the American flag flying from the roof. It is the building with the balcony and the columns. It is the building with the lawn and the trees. It is the building with the fence and the gate. It is the building with the flagpoles and the flag. It is the building with the windows and the doors. It is the building with the roof and the chimneys. It is the building with the stairs and the walkway. It is the building with the lights and the sign. It is the building with the name "The White House" written on it. It is the building that is the home of the President of the United States. \\
  \hline
 Rejection Sampling & The White House is the large, white building at the center of the image, directly in front of the greenhouse. The red house to the left is known as the Blair House, and it is located directly across the street from the White House. The greenhouse is the United States Department of Agriculture (USDA) Research Center, and the attractive heated pink place in the front is the USDA South Building.

The White House is the official residence and principal workplace of the president of the United States. It is located at 1600 Pennsylvania Avenue NW in Washington, D.C., and has been the residence of every U.S. president since John Adams in 1800. The Blair House is a historic house located at 1651 Pennsylvania Avenue NW, across the street from the White House. It serves as the official guest house for the president of the United States and is used for official entertaining and visiting dignitaries.

The USDA Research Center is a federal research facility that is part of the U.S. Department of Agriculture. It is located at 10301 Baltimore Avenue in College Park, Maryland, and is dedicated to research in agriculture, food, and natural resources. The USDA South Building is a federal office building that is part of the U.S. Department of Agriculture. It is located at 14th Street and Independence Avenue SW in Washington, D.C., and houses various USDA offices and programs.\\

\hline
PPO & The White House is the building in the center of the picture. It is the building with the flag flying from the roof. It is the building with the balcony and the columns. It is the building with the lawn and the trees. It is the building with the fence and the gate. It is the building with the flagpoles and the flag. It is the building with the windows and the doors. It is the building with the roof and the chimneys. It is the building with the columns and the arches. It is the building with the dome and the clock. It is the building with the flag and the stars. It is the building with the name "The White House" written on it. It is the building that is the home of the President of the United States.\\
\hline
  
  \end{tabular}
  \vspace{2pt} 
  \caption{Comparing generations obtained for a cherry-picked MT-Bench prompt from  Llama-2-7B trained based on different methods.}
  \label{tab:sample_mt_bench}
\end{table*}


\begin{table*}[h]
  \centering
  \renewcommand{\arraystretch}{1.5}
  \small
  \begin{tabular}{cp{13cm}}
    \hline
    \multicolumn{2}{p{16cm}}{\textbf{Prompt:} What are the five most common causes of stress among people?} \\
    \hline
    Policy & Response \\
    \hline
    Proposed Method & The five most common causes of stress among people are:

    Work: Job-related stress is one of the most common causes of stress, as it can involve long hours, tight deadlines, and demanding workloads.

    Money: Financial stress is another common cause of stress, as it can involve worries about debt, income, and financial stability.

    Relationships: Conflict with partners, family members, or friends can cause stress, as can the pressure to maintain healthy relationships.

    Health: Illness, chronic pain, or concerns about one's own health or the health of a loved one can cause stress.

    Personal problems: Events such as divorce, bereavement, or other personal difficulties can cause stress.

It's worth noting that these causes of stress can be interconnected, and that different people may experience stress for different reasons. It's important to take steps to manage stress and maintain good mental health. \\
     \hline
    SFT & The five most common causes of stress among people are:

1. Work or career-related issues
2. Money problems
3. Relationship difficulties
4. Health concerns
5. Personal problems

It is important to note that stress can also be caused by other factors such as personal beliefs, values, and experiences. Additionally, the perception of stress can vary from person to person, and what may be a source of stress for one person may not be a source of stress for another. \\

    \hline
  Original Annotation &The five most common causes of stress among people are:

1. Work or school
2. Money or financial concerns
3. Family responsibilities or relationships
4. Health concerns
5. Personal or emotional issues

It's important to note that the causes of stress can vary from person to person and may also be influenced by cultural and environmental factors. Additionally, it's possible for more than one cause to contribute to an individual's stress. \\
  \hline
 Rejection Sampling &Stress is a complex issue with many possible causes.  The most common causes of stress are often related to a person’s work or family life, and include:  financial difficulties, work-related stress, relationship problems, and health concerns.  Other common causes of stress include:  personal problems, such as a difficult relationship, divorce, or the death of a loved one; and life changes, such as moving, getting married, or having a baby.  Stress can also be caused by a person’s environment, including:  living in an unsafe neighborhood, being a victim of crime, or experiencing discrimination.\\

\hline
PPO & The five most common causes of stress among people are:

1. Work-related stress
2. Money problems
3. Relationship difficulties
4. Health issues
5. Personal problems

It is important to note that the causes of stress can vary from person to person and may be influenced by a variety of factors, including personal experiences, lifestyle, and cultural background.\\
\hline
  
  \end{tabular}
  \vspace{2pt} 
  \caption{Comparing generations obtained for a cherry-picked AlpacaEval prompt from  Llama-2-7B trained based on different methods.}
  \label{tab:sample_alpaca_eval}
\end{table*}


\begin{table*}[h]
  \centering
  \renewcommand{\arraystretch}{1.5}
  \small
  \begin{tabular}{cp{13cm}}
    \hline
    \multicolumn{2}{p{16cm}}{\textbf{Prompt:} Why does it feel like less effort to watch 3 - hour long TV episodes back to back than a film?} \\
    \hline
    Policy & Response \\
    \hline
    Proposed Method & There are a few reasons why it might feel like less effort to watch several hours of TV back to back than a single film:

1. Scope: TV shows often have a longer scope than films, with more time to develop characters, world-building, and plot. This can make the experience of watching several hours of TV feel more immersive and less like a single event.

2. Pacing: The pacing of TV shows is often more episodic, with each episode having a self-contained story that builds towards a larger arc over the course of the season. This can make it easier to follow the story and feel less overwhelmed by the amount of content.

3. Investment: Watching several hours of TV can also create a sense of investment in the characters and story, which can make it feel less like a one-time event and more like an ongoing experience.

4. Comfort: Watching TV can also be a comfortable and familiar activity, with a familiar format and rhythm that can be soothing and relaxing. This can make it easier to engage with the content for extended periods of time.

However, it's important to note that both films and TV shows can be enjoyable and engaging experiences, and the format preference is a personal one that depends on individual preferences and circumstances. Some people might prefer the immersive experience of a film, while others might find the longer format of TV more conducive to their lifestyle or entertainment preferences. Ultimately, the choice between films and TV shows is a matter of personal preference and can vary depending on the specific content, genre, and format preferences of the individual.\\

    \hline
  Original Annotation &There are several reasons why it may feel like less effort to watch three-hour long TV episodes back to back than a film. Firstly, the shorter duration of a film may make it seem more daunting to commit to watching it in one sitting, while the longer duration of a TV episode may feel more manageable. Additionally, the episodic format of TV shows often allows for natural breaks and pacing, which can make it easier to stay engaged and avoid feeling overwhelmed.

Another reason may be the structure of the content itself. TV shows tend to have more character development, plot twists, and subplots than films, which can make them feel more complex and engaging. On the other hand, films often have a more straightforward narrative structure and focus on a single story arc, which can make them feel more straightforward and easier to follow.

Finally, the cultural context in which we consume media may also play a role. With the rise of binge-watching and streaming services, it has become more common for people to watch multiple episodes of a TV show in one sitting. This has created a cultural expectation that may make it feel more normal and acceptable to watch longer periods of time, while the traditional film-going experience may still feel like a more significant investment of time and effort.

In summary, the combination of shorter duration, episodic structure, complex content, and cultural expectations may all contribute to the feeling that it is less effort to watch multiple TV episodes back to back than a film. However, it's important to note that this is a subjective experience and everyone's preferences and priorities may differ. \\
\hline
PPO & There are several possible reasons why it might feel like less effort to watch three-hour long TV episodes back to back than a film. One reason could be that the longer format of the TV episodes allows for more breaks and pauses, which can make the viewing experience feel less taxing. Additionally, the pacing of TV episodes can be more leisurely, which can make the content feel less overwhelming and easier to digest. Finally, the longer time commitment required for a TV series might make it feel like a more manageable task than a two-hour film, which can feel like a significant investment of time and attention. Ultimately, the perception of effort involved in watching content may be influenced by a variety of factors beyond the inherent length or format of the content itself.\\
\hline
  
  \end{tabular}
  \vspace{2pt} 
  \caption{Comparing generations obtained for a cherry-picked AlpacaEval prompt from  Llama-2-7B trained based on different methods.}
  \label{tab:sample_alpaca_eval2}
\end{table*}






\end{document}